\documentclass[11pt,a4paper]{article}
\usepackage[hyperref]{emnlp-ijcnlp-2019}
\usepackage{times}
\usepackage{latexsym}

\usepackage{makecell}
\usepackage{amsmath}
\usepackage{hyperref}
\usepackage{graphicx}
\usepackage{tcolorbox}
\usepackage{ulem}
\usepackage{tabularx}
\usepackage{soul}
\usepackage[T1]{fontenc}
\usepackage{commath}
\usepackage{booktabs}
\usepackage{todonotes}
\usepackage{float}

\usepackage{url}

\aclfinalcopy 

\title{End-to-End Entity Linking and Disambiguation leveraging Word and Knowledge Graph Embeddings}
\author{Rostislav Nedelchev \\
    University of Bonn \\
   Germany\\
   {\tt s6ronede@uni-bonn.de}\\\And
   Debanjan Chaudhuri \\
 University of Bonn \& Fraunhofer IAIS \\
  Germany\\
   {\tt s6dechau@uni-bonn.de} \\
   \AND 
   Jens Lehmann \\
   University of Bonn \& Fraunhofer IAIS \\
   Germany \\
   {\tt jens.lehmann@cs.uni-bonn.de} \\
   \vspace{6mm}
   \And
   Asja Fischer \\
   Ruhr University Bochum \\
   Germany\\
   {\tt asja.fischer@rub.de}\\
   \vspace{4mm}
   }
\date{}

\begin{document}

\maketitle

\begin{abstract}

Entity linking -- connecting entity mentions in a natural language utterance to knowledge graph 
 {(KG)} entities is a crucial step for question answering over KGs. {It is often based on measuring the string similarity between entity label and its mention in the question.} The relation {referred to} in the question can help {to} disambiguate between entities with the same label. 
{This can be misleading if an incorrect relation has been identified in the relation linking step}. 
 {However, an incorrect relation may still be semantically similar to the relation which the correct entity forms a triple with in the KG; which could be captured by a similarity of their KG embeddings. Based on this idea,} 
 we propose the first end-to-end neural network approach that {employs KG as well as} word embeddings 
 {to perform joint relation and entity classification of}
 simple questions while {implicitly performing} entity 
 {dismabiguation with the help of} a novel gating mechanism. {An e}mpirical evaluation {shows} that the proposed approach achieves {a performance comparable} to  state-of-the-art entity linking 
 {while requiring less} 
 {post-processing}.
  
\end{abstract}

\section{Introduction}

Question answering is a scientific discipline which aims at automatically answering questions posed by humans in natural language.  
Simple question answering over knowledge graphs is a well researched topic.
\cite{bordes2015large,yin2016simpleqaattentivecnn,mohammed2018strong,Petrochuk2018SimpleQuestionsNS}
A knowledge graph (KG) is a multi-relational graph which represents entities as nodes and relations between those entities as edges. Facts in a KG are stored in form of triples (\textit{h}, \textit{r}, \textit{t})  
where \textit{h} and \textit{t} denote the head (also called subject) and tail (also called object) entities, respectively, and \textit{r} denotes their relation.
A \textit{simple} question is a {natural language} question  {(NLQ)} that can be represented by a single (subject) entity  and a relation. Answering the query then corresponds to identifying the correct entity and relation given in 
{NLQ} and returning the object entity of {the} matching triple.   
For example, for the question \textit{Who is the producer of a beautiful mind?} {the corresponding KG fact is }
 (a beautiful mind, produced by, Brian Grazer) {and the} question answering system should be able to link to the correct entity "a beautiful mind" {(of type} movie) and the relation "produced by" in the KG to answer the question {by "Brian Grazer"}. 

The 
tasks of 
{identifying}
the {KG} entity {and relation} mentioned in the {NLQ} 
 {are} called entity {and relation} linking, {respectively}. 
The former {is often decomposed into} 
two sub-tasks, firstly detecting the span of the entity mention in the {NLQ}, secondly to connect {the identified mention} to a single entity in the KG, which is usually solved by comparing the entity mention {to} the names of the KG entities based on string similarity measures. This {becomes} particularly challenging {if} there 
{exist} more than one entity in the KG with the same label (name). However, the context provided for the  entity in the KG 
can be used 
{for disambiguation}. 
In our example, 
{correct relation linking would identify} the relation {"produced by"} {as being mentioned in the NLQ. Now one could make use of this information for entity linking by considering only entities which are connected to this specific relation. This would allow} 
to disambiguate and link to the movie "a beautiful mind" rather than the book.
This {procedure} is called soft disambiguation. 

However, relation linking
is {still} challenging since 
the
 number of relations in many KGs {is still large} (e.g.~6701 in {the} FB2M graph subset {of} the SimpleQuestions dataset \cite{bordes2015large}), while suffering from the problem of unbalanced classes \cite{xuquestion}.
Furthermore, {some} relations {may be} semantically similar, for example, 
{"fb:film.film.executive\_produced\_by"} and {"fb:film.film.produced\_by"}; and hence can be confusing for the relation linker. 
%
{Therefore, relation linking may end up predicting the wrong relation, which would negatively effect relation based entity disambiguation. To encounter this effect,}
it seems promising to leverage the relation specific information contained in the KG which is represented by the KG embedding of the relation. 
Semantically similar relations 
are closer to each other in KG embeddings vector space. So even if a model is not able to predict the
correct relation 
, the semantic information provided by 
KG embeddings can be employed to 
perform soft disambiguation of the entity candidates.

Based on these line of thoughts, 
we propose a novel end-to-end neural network 
model 
for answering simple questions from knowledge graphs 
, that incorporates both word and KG embeddings.
Specifically, the contributions of this paper are as follows:

\begin{itemize}
    \item 
    The proposal of a novel end-to-end model leveraging relatively simple architectures for entity and relation detection which is comparable to other state-of-the-art approaches for entity linking even without additional post-processing.
    \item The (to our knowledge) first investigation 
    of incorporating 
    KG embeddings for leveraging 
     KG structures for the end task of entity linking in an end-to-end manner.
    \item A novel gating mechanism incorporated in the end-to-end architecture which can implicitly perform entity disambiguation if required, improving overall entity linking performance. The final prediction is based on vector similarities, which along with the gate's output can be interpreted during prediction.
\end{itemize}




The rest of the paper is organized as follows: In Section 2, we summarize the related works on simple question answering. Sections 3 \& 4 provide the background and preliminaries important to this work. The overall approach and the architecture is explained in Section 5. In Section 6, we describe the experiment conditions. Evaluation results are discussed in Section 7. 
We do an ablation study and result analysis in section 8. Finally, we conclude and state the planned future works in Section 9.

\section{Related Work}

The SimpleQuestions dataset, as proposed by~\cite{bordes2015large} is the first large scale dataset for simple questions over Freebase. It consists of 108,442 questions split into train(70\%), valid(10\%),
and test(20\%). They also proposed an end-to-end architecture using memory networks along with the dataset.

The second end-to-end approach for simple question answering over Freebase was provided by~\cite{golub2016character}. They proposed a character LSTM based question encoder for encoding the question, a CNN based encoder for encoding the {KG} entity and relation, and finally an attention-based LSTM decoder for predicting an entity-relation pair given the question. {A similar ent-to-end approach was suggested by} \cite{lukovnikov2017neural}. 
 {It employs Gated Recurrent Unit (GRU) based encoders  that work on character and word level and in} 
addition {encode}
 the hierarchical types of relations and entities {to provide further information}.

{Furthermore, a growing set of}
modular architectures {was} proposed on the {SimpleQuestions} dataset. \cite{yin2016simpleqaattentivecnn} proposed a character-level CNN  
{for identifying entity mentions}
and a separate word-level CNN with attentive max-pooling to select knowledge graph tuples. \cite{yu2017improvedneural} utilized a hierarchical residual bidirectional LSTM for 
{for predicting a relation}, which is then used to re-rank the entity candidates. They replaced the topic entity in the question with a generic token <e> during relation prediction which helps in better distinguishing the relative position of each word compared to the entity.

\cite{dai2016cfo} proposed a conditional probabilistic framework with bidirectional GRUs that takes advantage of knowledge graph embeddings. \cite{mohammed2018strong}  suggested {to use a} combination of relatively simple, component-based approaches that {build on bidirectional GRUs, bidirectional LSTMs (BiLSTMs), and conditional random fields (CRFs)} 
{as well as on}
graph-based heuristics to select the most relevant entity {given a question from a candidate set. The resulting model} provides strong baselines for simple question-answering. 
More recently, \cite{huang2019knowledge} proposed an architecture based on {KG} embeddings,
\cite{Petrochuk2018SimpleQuestionsNS} proposed a technique 
{combining}
LSTM-CRF based entity detection {with Bi}LSTM based relation linking where they also replace the topic entity with generic tokens following~\cite{yu2017improvedneural}.

{Some other open-domain knowledge graphs are Wikidata\cite{vrandevcic2014wikidata} and DBpedia \cite{lehmann2015dbpedia}. In particular, there are two very recent efforts that provided adaptations of the SimpleQuestions dataset \cite{bordes2015large} to Wikidata \cite{diefenbach2017question} and DBpedia \cite{azmy2018farewell}. In addition, there is the Question Answering over Linked Data (QALD) \cite{unger20166th, usbeck20177th, usbeck20188th} series of challenges that use DBpedia as a knowledge base for QA.  }

\section{Background}

Knowledge graphs, word embeddings, and KG embeddings are concepts that are fundamental to this work. 
Embeddings 
{provide}
a numerical representation
{of words and KG entities/relations}
that facilitate {the incorporation of information provided by} 
KG and language {into} neural networks. 
{Their detailed description follows.}

\subsection{Knowledge Graphs}

In this work, a KG is a network of real world entities that are connected to 
{each other
}
by means of relations. Those entities and their relations are represented as nodes and edges respectively in a multi-relational, directed graph. Knowledge graphs consist of ordered triples, also known as facts, of the form $(s, r, o)$, where {$s$ and  $o$} are two entities connected by the  {relation $r$}.
Revisiting the 
example from the introduction, 
{the corresponding fact would be given by}
("a\:beautiful\:mind"
"produced\:by", "Brian\:Grazer").

\subsection{Knowledge Graph Embeddings}

{Knowledge graphs are 
data structures }
that lack a default numerical representation that allows their straightforward application in a  standard machine learning context.
Statistical relation learning therefore relies beside other approaches on latent feature models for making prediction over KGs. These latent features usually correspond to embedding vectors of the KG entities and relations. Given the embeddings of the entities and relations of a fact $(h,r,t)$, a score function outputs the a value indicating the probability that the fact is existing in the KG.
%
In this paper, we use  TransE \cite{bordes2013transe} {to learn the KG embeddings used by our model}. 
Let the embedding vector of subject and object entity be given by $\vec h$ and $\vec t$ respectively, and that of the relation by a vector $\vec r$, then the score function of TransE is given by
$f(h,r,t) = -\| \vec h+ \vec r- \vec t \|$. 


\subsection{Word Embeddings}

{In recent times, various approaches were proposed that embed words on a vector space \cite{bengio2003neural, collobert2008unified}. These methods create representations for each word as a vector of floating-point numbers. Words whose vectors are close to each other are demonstrated to be semantically similar or related. Especially, two works \cite{mikolov2013efficient, mikolov2013distributed} showed the high potential of word embeddings in natural language processing (NLP) problems.}

{In this work, we use the GloVe \cite{pennington2014glove} vector embeddings. The method aims to create a $d$-dimensional vector representation of each word in the vocabulary, $w \in V_w$, and the context of each occurrence of a word in the corpus, $c \in V_C$ so that:}
$$\vec{w} \cdot \vec{c} + b_w + b_c = \log(\# (w,c)) \quad \forall (w,c) \in D 
$$
\noindent {where $b_w$ and $b_c$ are biases 
that are learned together with $\vec{w}$ and $\vec{c}$ and $\# (w,c)$ is the word occurrence count of word $w$ in context $c$.}

\noindent $D$ being the documents from which the corpora is extracted from.

\section{Preliminaries}

We {employ} two kinds of embeddings in the proposed model namely word embeddings and KG embeddings which are defined in the previous section. 


{For matching a question to the entities and relations of a KG, likely candidates are first selected in an reprocessing step to reduce the enormous number of candidates in the KG. This is described in the following sections.}

\subsection{Entity Candidates Generation} \label{sub:candgen}



We first start with a simple {language based} candidate generation process {selecting potential candidate entities for a given question. That is,} given a question we generate candidates by matching the tf-idf vectors of the query with that of the entity labels of all the entities in the knowledge graph, {resulting in} a list of $n$ entity candidates $e_c^1, e_c^2.. e_c^n$ for {a given} question. 
This list is then re-ranked based on the tf-idf similarity score, whether the candidate label is present in the question, number of relation the candidate is connected to in the KG and whether it has a direct mapping to wikipedia or not. This is done to give importance to important entities (defined by connectivity in the KG) following previous works \cite{mohammed2018strong} \cite{Petrochuk2018SimpleQuestionsNS}.

{
\subsection{Relation Candidate Generation}
To generate a set of entity specific relation candidates, for each entity $e_c^j$ in the entity candidate set we extract the list of relations connected to this candidate at a 1-hop distance in the Freebase Knowledge Graph. 
For the $j^{th}$ entity candidate $e^j_c$, the relation candidates are $r_j^1...r_j^n$.
}

\section{Model Description}

\begin{figure*}[ht]
    \centering
    \vspace*{-0.1cm}
    \includegraphics[width=0.65\textwidth]{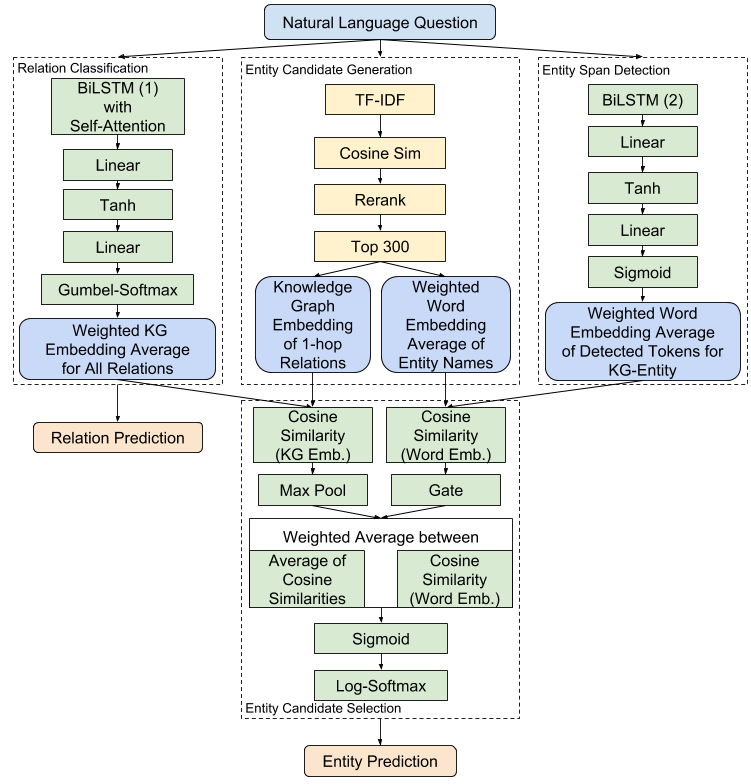}
    \caption{Architecture}
    \label{fig:Architecture}
    \vspace*{-0.1cm}
\end{figure*}

The proposed neural network model is quintessentially composed of {three parts} as visualized in \ref{fig:Architecture}:

\begin{enumerate}
    \item A {word-embedding} based \textbf{entity span detection model}, which selects the probable 
    {words of an}
    entity in a natural language question, {represented by a bi-LSTM}.
    \item An 
     {word-embedding}  based \textbf{relation {prediction} 
    model} which links the 
    the question to one of the relations in the knowledge graph, {represented by a bi-LSTM with self-attention}.
    \item {An \textbf{entity prediction model} which takes the predictions of the previous two submodels into  account and employs a sentinel gating mechanism that performs disambiguation based on similarity measures.}
\end{enumerate}
{The different model parts and the training objective of the resulting model will be described in more detail in the following.}

\subsection{Entity Span Detection}
\label{sec:e_span_detect}



The span detection module is inspired by \cite{mohammed2018strong}. The given question is firstly passed through a bi-directional LSTM. 
{Its output} hidden states 
are then passed through a fully connected layer and a {sigmoid ($\sigma$)} activation function which {outputs the probability of} the word at time-step $t$ {corresponding to} an entity (or not). Mathematically, {this can be described as} 
\begin{equation}
\begin{split}
    & h_t = f_{span}(x_t) \enspace ;
     o_t = \sigma(Wh_t)
\end{split}
\end{equation}
where, $o_t$ is the output 
{probability,}
$W$ the weight {matrix of the fully connected layer,} and $h_t$ the 
hidden state {vector from} the applying the bi-directional LSTM $f_{span}$ on the input question $x_t$.  We use I-O encoding for the output for training. 

\subsection{Relation Prediction}
\label{sec:rel_detec}



For relation prediction, the question is passed into a self-attention {based} bi-directional LSTM {which was} inspired by \cite{zhou2016attentionrl}.
The {attention} weighted hidden states are then fed into a fully-connected {classification} layer {outputting a probability over the $r_n$ relations in the knowledge graph}. 

\begin{equation}
    y_r = tanh(W_r(\alpha_r * f_{rel}(x_t))) 
\end{equation}
$W_r$ being a set of model parameters, $\alpha_r$ the self-attention weights and $f_{rel}$ is the Bi-LSTM function which produces a response for every time-step $t$ for the input query $x_t$. 

\subsection{Entity prediction}
\subsubsection{{Word-based} Entity Candidate Selection.}
\label{sec:W_ent_cad_sel}

{With the help of the entity span identified by the span detection submodule described in \ref{sec:e_span_detect}, the questions are now compared to the}
entity candidates 
based on vector similarity methods. {More specifically, the word embedding of each word of the question is multiplied with corresponding} output {probability} from the entity-span detection model  
{leading to an "entity-weighted" word representation}
\begin{equation}
    e_t = o_t * w_t^{emb} 
\end{equation}
where, $w_t^{emb}$ denotes the word embedding of the $t$-th word in the question and $o_t$ is the sigmoid output from \ref{sec:e_span_detect}. We {then} take a simple average {of the entity-weighted  representations of all words of the questions to yield the entity embedding of the question $e_q^{emb}$.}

Similarly, the entity candidates 
{$e_c^1, e_c^2.. e_c^n$ generated in the preprocessing step described in \ref{sub:candgen} are represented by the word embeddings of their labels $e^{emb1}_c$, $e^{emb2}_c$.. $e^{embn}_c $.}
{If a label consists of multiple words, the word embeddings are averaged to yield a singe representation.}
 Finally, {to compute the similarity between a question and an entity candidate}, the cosine between the
 question embedding and the entity embedding is estimated.
 For the $j^{th}$ candidate, {that is} 
\begin{equation}
    sim_c^j = cos(e_q^{emb}, e^{embj}_c) \enspace
\end{equation}
{and the vector $sim_c=(sim_c^1, \dots sim_c^n)$ represents the word based similarity of the question to all entity candidates.}


%
%

\subsubsection{
{KG-based Entity Candidate Selection.}
}
\label{sec:KG_ent_cad_sel}
{To leverage the relational information encoded in the KG, we firstly take the logits over $r_n$ from the relation prediction model and draw a categorical representation using gumbel softmax \cite{jang2016categorical}. This representation is multiplied with the KG embeddings over $r_n$ to get a KG embedding based representation of the query $r^{emb}_q$.  
This relation specific representation  is then compared against the full relation candidate set of each candidate entity, where each candidate relation is as well represented by its KG embedding.}
To match 
{the relation specific question representation to relation candidates for a given entity,}
{we estimate the} cosine similarity of the {corresponding} KG embeddings followed by a max-pooling operation over all the candidate relations of an entity which produces an {entity specific} similarity metric $sim_{kg}^j$, {which indicates the degree of matching between the question and an entity candidate from a KG perspective, which specifically takes relation information into account}. Mathematically, for the $j$-th entity candidate, {let} the embedding {of} the $k$-th {relation} candidate $r_j^k$ {be} denoted by $r^{embk}_j$. {The KG based similarity $sim_{kg}^j$ between the question and the $j$-th entity then given by}
\begin{equation}
\begin{split}
    sim_{kg}^j = maxpool( & cos(r^{emb}_q, r^{emb1}_j), \\
                          & cos(r^{emb}_q, r^{emb2}_j) \\
                          & .. \\
                          & cos(r^{emb}_q, r^{embk}_j)) \enspace,
\end{split}
\end{equation}
{and the vector $sim_{kg}=(sim_{kg}^1, \dots sim_{kg}^n)$ represents the KG based similarity of the question to all entity candidates.}




\subsubsection{{Disambiguation and final prediction.}}

{The final entity prediction is based on the word- and KG-based similarity measures $sim_{c}$ and $sim_{kg}$ 
}
{First, for disambiguation, the word based similarity vector $sim_c$ is}
passed into a gating mechanism 
%
\begin{equation}
    g_{amb} = W_gsim_c \enspace
\end{equation}
{with} 
$W_g \in \mathbf{R}^{nX1}$, 
{which aims at estimating if there is more than one single likely candidate in the entity candidate set based on word similarity. If so, the KG based similarity $sim_{kg}$ should also be taken into account, which is done by averaging $sim_{c}$ and $sim_{kg}$ 
and predicting the final entity candidate by}
\begin{equation}
\begin{split}
    y^e_p = 
    \sigma( & g_{amb} \space * mean(sim_{kg}, sim_c) \\
             & + (1-g_{amb})*sim_c)
\end{split}
\end{equation}
{Note that,}
$y^e_p$ $\in$ $\mathbf{R}^n$ are the logits over
{the set of candidate entities, from which the entity with the highest probability can be picked.} 
During inferencing, we perform an additional step for ensuring the entity and relation predicted from the model forms a pair in the KG. In order to achieve that, we take the top 5 probable relation from the relation linker and choose the one which is connected to the predicted entity at 1-hop.

\section{{Training}
}

\subsection{{Training objective}} 
The model is trained based on a multi-task objective, where the total loss is the sum of the losses from the entity span detection, relation detection, entity candidate prediction, and disambiguation. The individual loss function are given below\footnote{$y$ is used to denote the true label for all tasks here}.


\noindent {The loss function for the entity span detection model is the average binary cross entropy $ L_{span}  $ over the words of the input question, with}
\begin{equation}
\begin{split}
    & L_{span} = \frac{1}{T}\sum_{t=1}^T {l_t} \\
    & l_t = - [ y_t \cdot \log \sigma(o_t) + \\
    & \qquad\quad (1 - y_t) \cdot \log (1 - \sigma(o_t)) ] \enspace,
\end{split}
\end{equation}
where $y_t$ is the label 
denoting if the $t$-th word belongs to the entity span or not.
For relation prediction, a weighted cross-entropy loss $L_{rel}$ is used (where the weights are given by the relative ratio of relations in the training set having the same class as the sample)
\noindent {and for  entity prediction a vanilla  cross-entropy loss $L_{ent}$, which depends on the parameters of all sub-models.}
Furthermore, an additional cross entropy loss function  $L_{amb}$ is used to train the gating function. 
%
Last but not least, we add an regularization term for soft-parameter sharing following \cite{duong2015low} resulting in a total loss given by
\begin{equation}
\begin{split}
    L = & L_{span} + L_{rel} + L_{ent} + L_{amb} \\
        & + ||W_{span}^1 - W_{rel}^1||^2 \enspace,
\end{split}
\end{equation}
where, $W_{span}^1$ and $W_{rel}^1$ are the hidden layer weights of the entity span detection and relation detection module. 
 Given $L$, all  parameters of the model are {jointly} trained in an end-to-end manner. 





\subsection{{Training details}}

We use the pre-processed data {and word-embeddings} provided by \cite{mohammed2018strong} to train our models. 
{To obtain} KG embeddings, we train TransE \cite{bordes2013transe} on the provided Freebase KG of 2 million entities. The {size of the} word embedding vectors is 300, and that of the KG embeddings is 50. The KG embeddings are kept fixed but the word embeddings are {fine-tuned} during optimization. For training the disambiguation gate $g_{amb}$ we use {a label of} 1 if the correct entity label is present more than one times in the entity candidates, {and label of} 0 otherwise. 

For training, a batch-size of 100 is used and the model is trained for 100 epochs. We save the model with the best validation accuracy {of} entity prediction and evaluate {it on the test set.} We {apply Adam \cite{kingma2014adam} for optimization with a} learning rate of 1e-4. 
The {size of the hidden layer} of both the entity span and relation prediction Bi-LSTM is {set to} 300. The training process is {conducted} on a GPU with 3072 CUDA cores and a VRAM of 12GB.  

\section{Evaluation}

In this section, we compare our model with other state-of-the-art entity linking and question-answering systems, both end-to-end and modular approaches.
{Resources are provided as supplementary materials to this paper that allow the reader to reproduce the final results reported in this section.}

\subsection{Entity-linking}

We compare our end-to-end entity-linking accuracy with other systems whose results are published, on the test set. The results are summarized in Table \ref{tab:ent_linking}. The number of candidates $n$ in the entity candidate set is varied from 100 to 300. The percentage of examples for which the correct entity candidate is present in the candidate set is  reported in parenthesis. If the percentage is higher, model performance also increases. The model evaluated over wthe largest entity candidate set (i.e. $n$ = 300) gives the best performance, which is significantly better than the BiLSTM based model \cite{mohammed2018strong} (13.60\% additional accuracy) and the Attentive CNN model (5\% additional accuracy). It must be noted that our model cannot be compared directly to the one from \cite{mohammed2018strong} because they don't use any candidate information for entity-linking, they do it during final question-answering as a post-processing step. The BiLSTM based model in combination with a n-gram based entity matching and relation based re-ranking suggested by \cite{yu2017improvedneural} is better than our proposed model by 0.40\%.

\begin{table}[ht]
\vspace{-0.1cm}
    \centering 
     \begin{tabular}{ l | c}
     \toprule
     \textbf{\thead{Model}} & \textbf{\thead{Accuracy\\(\% of Cand. Present)}} \\
     \hline
     \makecell{BiLSTM \\ \cite{mohammed2018strong}} & 65.00 (-) \\
     \hline
     \makecell{Attentive CNN \\ \cite{yin2016simpleqaattentivecnn}} & 73.60 (-) \\
     \hline
     \makecell{BiLSTM \& Entity-reranking \\ \cite{yu2017improvedneural}} &	\textbf{79.00} (-)\\
      \midrule
     Proposed model (n=100)	& 77.80(92.07) \\
     Proposed model (n=200)	& 78.35(94.34)\\
     Proposed model (n=300)	& \textbf{78.60}(95.49) \\

     \bottomrule
    
    \end{tabular}
       \vspace{-.1cm}
     
    \caption{Entity Linking Accuracy.}
    \label{tab:ent_linking}
    
\end{table}
\vspace{-0.5cm}

\subsection{Question Answering}

The 
final metric for simple QAKG is defined by the number of correct entities and relations predicted by a given model. We are comparing the performance of our system with that of both end-to-end methods and modular approaches in Table \ref{tab:qa}.
The results show that the proposed architecture outperforms the state-of-the-art NN based model (GRU based) \cite{lukovnikov2017neural} by 2.0 \%, and shows a  performance competitive to simple modular  baseline approaches like  \cite{mohammed2018strong} and the KG embedding based approaches KEQA proposed by \cite{huang2019knowledge}. However, the best state-of-the-art approach on QAKG \cite{yu2017improvedneural} outperforms ours model by 5.50\%.  It should be noted here that although our entity-span detection and relation linking accuracy (82.01 \%) is better than that of the model proposed by \cite{mohammed2018strong}, the 
final question answering performance is worse by 1.7 \%. This can be explained by the fact that their approach builds on additional string-matching heuristics along with the scores from the different models to re-rank the predicted entities and relations.

\begin{table*}[]
     \centering 
      \vspace{-.1cm}
     \begin{tabular}{ l | c | c}
   
     \toprule
     \textbf{Approach} & \textbf{Model} & \textbf{   Accuracy(FB2M)  } \\
     \hline
     End-to-End NN & Memory NN \cite{bordes2015large} & 61.60\\
     & Attn. LSTM \cite{golub2016character} & 70.90 \\
     & GRU based \cite{lukovnikov2017neural} &	{71.20}\\
           \cmidrule{2-3}
     &Proposed model (n=100)	& 72.29 \\
     &Proposed model (n=200)	& 72.84\\
     &Proposed model (n=300)	& \textbf{73.20} \\
     \midrule
     Modular & BiLSTM \& BiGRU \cite{mohammed2018strong} & 74.90 \\
     & KEQA \cite{huang2019knowledge} & 75.40 \\
     & CFO \cite{dai2016cfo} & 75.70 \\
     & CNN \& Attn. CNN \&  BiLSTM-CRF \cite{yin2016simpleqaattentivecnn} & 76.40 \\
     & BiLSTM-CRF \& BiLSTM \cite{Petrochuk2018SimpleQuestionsNS} & 78.10 \\
     & BiLSTM \& Entity-reranking \cite{yu2017improvedneural} & \textbf{78.70} \\
     \bottomrule
    \end{tabular}
    \vspace{-0.2cm}
    \caption{Question Answering Accuracy.}
    \label{tab:qa}
\end{table*}

\begin{table}[ht]
     \begin{tabular}{ l | c}
     \toprule
     \textbf{\thead{Approach}} & \textbf{\thead{Entity-linking \\ Accuracy}} \\
     \hline
    {Removing $L_{rel}$ from total loss} & 67.55 \\
    \hline
    \makecell{Removing gating mechanism $g_{amb}$} & 74.63 \\
    \hline
    \makecell{Removing soft-loss from total loss} & 78.17 \\
    \hline
    {Without re-ranking candidates} & 77.56 \\
    \hline
    Our Best Model (n=300) & 78.60 \\
    \bottomrule
    \end{tabular}
     
    \caption{Ablation Study}
    \label{tab:ablat}
    \vspace*{-0.1cm}
\end{table}

\section{Discussion}




\subsection{Ablation Study}

Finally, we do an ablation study where we remove some parts of the proposed model and observe the performance of entity linking for $n$ = 300. The results are in \ref{tab:ablat}. As observed, the entity-linking accuracy from not training the relation linker are at par with \cite{mohammed2018strong} in Table \ref{tab:ent_linking}. The gating mechanism adds 3.97 \%, 
because doing only a mean from the entity and relation prediction similarity scores would add in extra information overhead for the candidate selection for wrongly classified relation. The proposed soft-loss aids in 0.43 \% increase in entity-linking accuracy and the candidate re-ranking improves it by 1.04\%. 


\subsection{Quantitative and Error Analysis} \label{sec:heuristics}

We do a quantitative analysis from the results of our best model with $n$=300. Percentage of questions with soft-disambiguity is 21.1 \% and with hard-ambiguity is 18.51 \%. Our model is able to predict 84.81 \% of correct entity candidates for soft-disambiguation cases, out of which 75.02 \% of times the correct relation was identified and 9.78 \% the model predicted the wrong relation but the correct candidate is picked using our proposed KG embeddings based method; which proves that our intuition for using KG embeddings for the final task can be beneficial. For hard-ambiguity cases, the model was able to predict the correct candidate with an accuracy of 35.66 \% (1432 out of 4015 cases), out of which the model predicted wrong relations 4.4 \% of cases. But, it should be noted that there are no explicit linguistic signals to solve hard-disambiguity, following previous works we are predicting these cases based solely on candidate importance. 

The model is able to predict the correct candidate 97.70 \% of the times for cases where no disambiguation is required. Out of the 440 such wrongly classified candidates, 165 cases are because the true entity and correct relation are not connected in the KG at 1-hop, 162 because the entity span detector was not able to predict the correct span and the rest for wrong prediction in the disambiguate gating mechanism.

In general, some cases where the entity-span detector has failed to identify the correct entity is in table \ref{tab:span_err}. In some of these cases, there are more than 1 entity in the question. Hence, it is difficult for the entity span detector to detect the correct entity.

\begin{table}
     \begin{tabular}{p{70mm}}
     \toprule
    what 's a \colorbox{green!20}{rocket} that has been \colorbox{blue!20}{flown} \\
    \hline
    who is a \colorbox{green!20}{swedish} \colorbox{blue!20}{composer}  \\
    \hline
    what 's the name of an \colorbox{green!20}{environmental disaster} in \colorbox{blue!20}{italy}  \\
    \hline
    which \colorbox{green!20}{korean air} flight was in an \colorbox{blue!20}{accident} \\
    \bottomrule
    \end{tabular}
    \caption{Span Detection Error. Green - correct span, blue - detected span.}
    \label{tab:span_err}
    \vspace*{-0.1cm}
\end{table}

For the final question-answering task, as mentioned previously, although the end-to-end accuracy for \cite{mohammed2018strong} is better than ours', but the task of question answering is particularly challenging in this case because we don't use any scores from string matching based methods such as Levenshtein distance for entity linking as done as an additional post-processing step by \cite{mohammed2018strong}, especially in cases where the entity candidates and the entity mention in the question consists of out-of-vocabulary words. Also, for some cases, it is challenging to disambiguate between the predicted relations because there are no explicit linguistic signals available. To exemplify, let us consider the question \textit{ what county is sandy balls near ?}. The predicted relation relation for this question by our model is {"fb:location.location.containedby"} while the true relation in the dataset is {"fb:travel.tourist\_attraction.near\_travel\_destination"}.

\section{Conclusion and Future Work}

In this paper, we have proposed an end-to-end model for entity linking, leveraging KG embeddings along with word embeddings banking on relatively simple architectures for entity and relation detection. As reported, the proposed architecture performs better than other end-to-end models but modular architectures demonstrates better question answering performance. However, the purpose of this paper was to integrate KG and word embeddings in a single, end-to-end model for entity linking. Moreover, since the final prediction model is based on similarity scores, the final prediction (and gating) can be easily interpreted following equations 4, 5, 6 and 7. 

Error analysis suggest that the model can gain from better entity span detection. As a future work we will experiment by integrating CRF-biLSTM for span-detection and also with more recent NLP models like BERT. The model will also improve with better relation linking and better handling of out-of-vocabulary words. We would also like to integrate more recent state-of-the-art KG embedding models \cite{dettmers2018conve,schlichtkrull2018modeling}, which can capture better relation semantics in the architecture as a future work.



\bibliography{emnlp-ijcnlp-2019}
\bibliographystyle{acl_natbib}


\end{document}